\documentclass[conference]{ieeeconf}

\usepackage{copyright}

\IEEEoverridecommandlockouts

\usepackage{graphicx}
\usepackage{booktabs}
\usepackage{pifont}

\usepackage{amsmath}
\DeclareMathOperator{\atantwo}{atan2}

\usepackage{amssymb}
\usepackage{mathtools}
\usepackage{physics}
\usepackage{xfrac}
\usepackage{amsfonts}

\usepackage{pifont}

\usepackage{threeparttable}
\usepackage{subcaption}
\usepackage{tabularx}

\DeclarePairedDelimiterX\set[1]\lbrace\rbrace{#1}

\usepackage{cleveref}
\crefname{table}{Tab.}{Tabs.}
\crefname{figure}{Fig.}{Figs.}
\crefname{section}{Sec.}{Secs.}
\crefname{equation}{Eq.}{Eqs.}

\usepackage[acronym]{glossaries}
\newacronym{cctv}{CCTV}{closed-circuit television}
\newacronym{pd}{PD}{proportional derivative}
\newacronym{mlp}{MLP}{Multi-Layer Perceptron}
\newacronym{rmse}{RMSE}{Root Mean Square Error}
\newacronym{cnn}{CNN}{Convolutional Neural Network}
\newacronym{miou}{mIoU}{mean Intersection over Union}
\newacronym{rte}{RTE}{Relative Translation Error}
\newacronym{rre}{RRE}{Relative Rotation Error}
\newacronym{ict}{ICT}{Information Communications Technology}
\newacronym{iot}{IOT}{Internet of Things}
\newacronym{pbvs}{PBVS}{Position-Based Visual Servoing}
\newacronym{ibvs}{IBVS}{Image-Based Visual Servoing}
\newacronym{fov}{FOV}{Field-of-View} 
\newacronym{dl}{DL}{Deep Learning} 
\newacronym{fc}{FC}{Fully-Connected} 
\newacronym{mse}{MSE}{Mean Square Error}

\usepackage{cuted}

\usepackage{tikz}
\usetikzlibrary{fit,calc}

\usepackage{siunitx}
\usepackage{xcolor}

\title{\huge \bf
Visual Servoing on Wheels: Robust Robot Orientation Estimation in Remote Viewpoint Control
}
\author{Luke Robinson$^{*}$, Daniele De Martini, Matthew Gadd, and Paul Newman
\\
Mobile Robotics Group (MRG), University of Oxford\\\texttt{\{lrobinson,daniele,mattgadd,pnewman\}@robots.ox.ac.uk}
\thanks{$^{*}$Corresponding author.
Thanks to the EPSRC Programme Grant ``From Sensing to Collaboration'' (EP/V000748/1).
}%
}

\begin{document}

\maketitle

\copyrightnotice

\begin{abstract}
This work proposes a fast deployment pipeline for visually-servoed robots which does not assume anything about either the robot -- e.g. sizes, colour or the presence of markers -- or the deployment environment.
Specifically, we apply a learning based approach to reliably estimate the pose of a robot in the image frame of a 2D camera upon which a visual servoing control system can be deployed.
To alleviate the time-consuming process of labelling image data, we propose a weakly supervised pipeline that can produce a vast amount of data in a small amount of time.
We evaluate our approach on a dataset of remote camera images captured in various indoor environments demonstrating high tracking performances when integrated into a fully-autonomous pipeline with a simple controller.
With this, we then analyse the data requirement of our approach, showing how it is possible to deploy a new robot in a new environment in less than \SI{30}{\minute}.
\end{abstract}


\glsresetall

\section{Introduction}%
\label{sec:introduction}

Visual detection has become an increasingly popular robot-control method.
In particular, visual servoing has been studied as a way to control a robot using a single remote camera, tracking features in the camera image to estimate the robot's pose, and then using this pose to generate control signals for the robot's motion.
Combined with the possibility to offload computing power, a robot will be not much more than a drive-by-wire apparatus capable of receiving actuator commands in a shared, smart \gls{ict} infrastructure.

Although much work has been done in visual servoing to obtain reliable control architectures, deploying such methods in real-world scenarios is still an open problem.
Indeed, traditional visual servoing methods often require markers or other visual cues to be attached to the robot, which can limit their applicability in certain scenarios \cite{liang2015adaptive,liang2020purely}.

While object detection and tracking pipelines are mature and already used in commercial applications, estimating objects' orientation in the image plane is still largely unexplored.
In this paper, we present a fast deployment framework for visual servoing applications supported by a learnt orientation estimation in the image plane without requiring any markers on the robot.
Specifically, we train a neural network to estimate the robot's pose by creating vast amounts of data through random walks of the robot in the camera frame.
All training is done directly on data in the deployment environment with a total set-up time of fewer than \SI{30}{\minute}.


\begin{figure}[!t]
\centering
\includegraphics[width=\columnwidth]{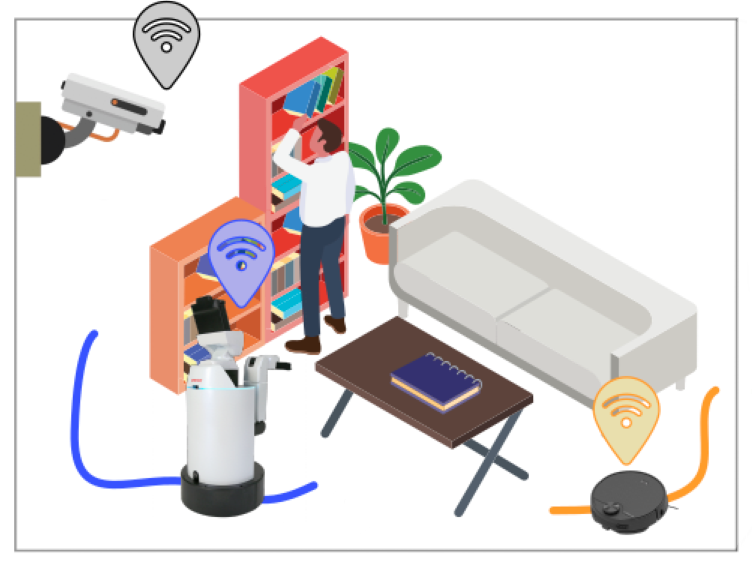}
\caption{In productive indoor scenes such as warehouses and working offices, clutter and activity can create a confusing background for visual servoing perception systems, making it difficult to accurately estimate the robot's pose.
Nevertheless, an easy and fast setup is imperative for the wide adoption of these systems.
To overcome these challenges, we present a lightweight orientation estimator and a training procedure for quickly producing models which are expert in particular workspaces with almost no manual labelling.
We demonstrate the benefit of this much-improved orientation estimation within the context of off-board \acrshort{cctv} control, showing that our system bolsters the precision and efficiency of robot autonomy.
}
\label{fig:front_page}
\end{figure}

We demonstrate the effectiveness of our approach through experiments on a dataset of remote camera images captured in various indoor environments.
Our results show that our approach can accurately estimate the robot's pose in real-time, even in cluttered environments, producing low tracking errors when integrated into a fully-autonomous pipeline.

To summarise, our contributions are:
\begin{itemize}
\item A framework for fast deployment of a robot in a visual servoing context;
\item A lightweight orientation estimation network capable of \SI{360}{\degree} angle estimation and a training procedure for weak supervision;
\item A thorough evaluation of the pipeline in several indoor settings, with ablation studies on data requirements;
\item The integration of the proposed pipeline in a full-autonomy test.
\end{itemize}

Indeed, we present our orientation estimator within an off-board system which is totally calibration-less and lacking fiducial markers such as the use of floor markers in~\cite{shim2016mobile}, and using only a monocular remote viewpoint as opposed to stereo~\cite{flogel2022infrastructure}.

\section{Related Work}%
\label{sec:literature}

Visual servoing is the act of controlling a robot through the use of visual sensor data, a form of control that has shown consistent development throughout the last forty years \cite{agin1980computer}.
This continued development is largely due to its ability to detect information about a scene without physical contact and in a form that people can easily understand.

The field of visual servoing is traditionally split into two areas -- applications in which the camera is mounted on the robot (robot-in-hand) as opposed to applications where the camera is statically mounted in the environment and is viewing the robot remotely (robot-to-hand) \cite{flandin2000eye}.
As this work concerns robot-to-hand research, this review will focus on the latter.

Visual servoing work can also be distinguished by the space the robot is controlled within, with the two main categories being \gls{pbvs} and \gls{ibvs}.
\Gls{pbvs} methods \cite{lippiello2004visual, janbi2010kalman, xue2021performance} use the camera parameters and 3D knowledge of the robot and environment to extract the robot's position in world coordinates from the image frame.
This has the benefit of recovering the full 3D pose of the robot so that more traditional, well-developed control methods can be used.
Still, \gls{pbvs} is sensitive to the camera parameters \cite{shademan2005sensitivity} and the separation of controller and image processing does not guarantee that the control signals given will keep the robot in the image \gls{fov}.
In \gls{ibvs} \cite{tsakalis2019adaptive}, instead, both sensing and control happen in the image space, alleviating these disadvantages but imposing the restriction of stating the desired pose of the robot in terms of an image state.
This is challenging as neither depth information nor 3D model of the robot is available.
Techniques that blend \gls{pbvs} and \gls{ibvs} are known as hybrid visual servoing \cite{malis1999hybrid}, some of which have exploited modern \gls{dl} to learn the robot control signals directly from the raw images \cite{machkour2022classical}.
These latter methods, however, are mainly limited in use to robot-in-hand systems.

Our work is situated within the \gls{ibvs} field because we aim to deploy a visual servoing system as fast as possible, and as such, we allow ourselves no time-consuming calibration procedures and no prior knowledge of robot or environment geometry.
In doing so, we aim to recover the position and orientation of the robot and control it in purely image space.

An early innovator for \gls{ibvs} for non-holonomic ground vehicles was \cite{dixon2001adaptive}, which proposed an approach to track and control a wheeled robot using a statically mounted camera with no knowledge of the specific motion model, camera intrinsic, or height of the camera above the ground.
However, the image plane needed to be parallel to the ground and the robot had two LEDs that could be easily detected for position and orientation in the frame.
Since then, various work has been done on the problem \cite{wang2010dynamic, liang2015adaptive, yang2018adaptive} leading to the current state-of-the-art \cite{liang2020purely}, where all restrictions on the pose of the camera are removed, allowing a completely uncalibrated camera and robot model to follow a track on the image plane without a learning phase.
However, the focus of~\cite{liang2020purely} was on the control policy and stability analysis, and the robot was still required to be marked with specific, predetermined feature points for position and orientation estimation.

This paper is an extension to this theme, further relaxing any required calibration -- i.e. not placing known markers on the robots.
Indeed, we propose a \textit{learned} object detector and a novel orientation estimator to recover the robot's pose, both of which are trained extremely efficiently, with minimal manual collection, annotation or effort.
Indeed, our networks are trained directly on the deployment robot and environment, with a limited dataset dictated by the fast-deployment requirements.

While much work has been done on object detection, leading to very mature pipelines \cite{diwan2022object}, less work has been done on orientation detection.
Closest to our work, \cite{ruiz2018fine} represents a 3D rotation in the range \SIrange{-99}{99}{\degree} as three classification problems -- carried out by three \gls{fc} heads -- then projected into a single scalar.
We took inspiration from \cite{ruiz2018fine} and extended it to our specific problem, where we need to estimate a single angle between \SI{0}{\degree} and \SI{360}{\degree}.

\begin{figure}
\centering
\includegraphics[width=\linewidth]{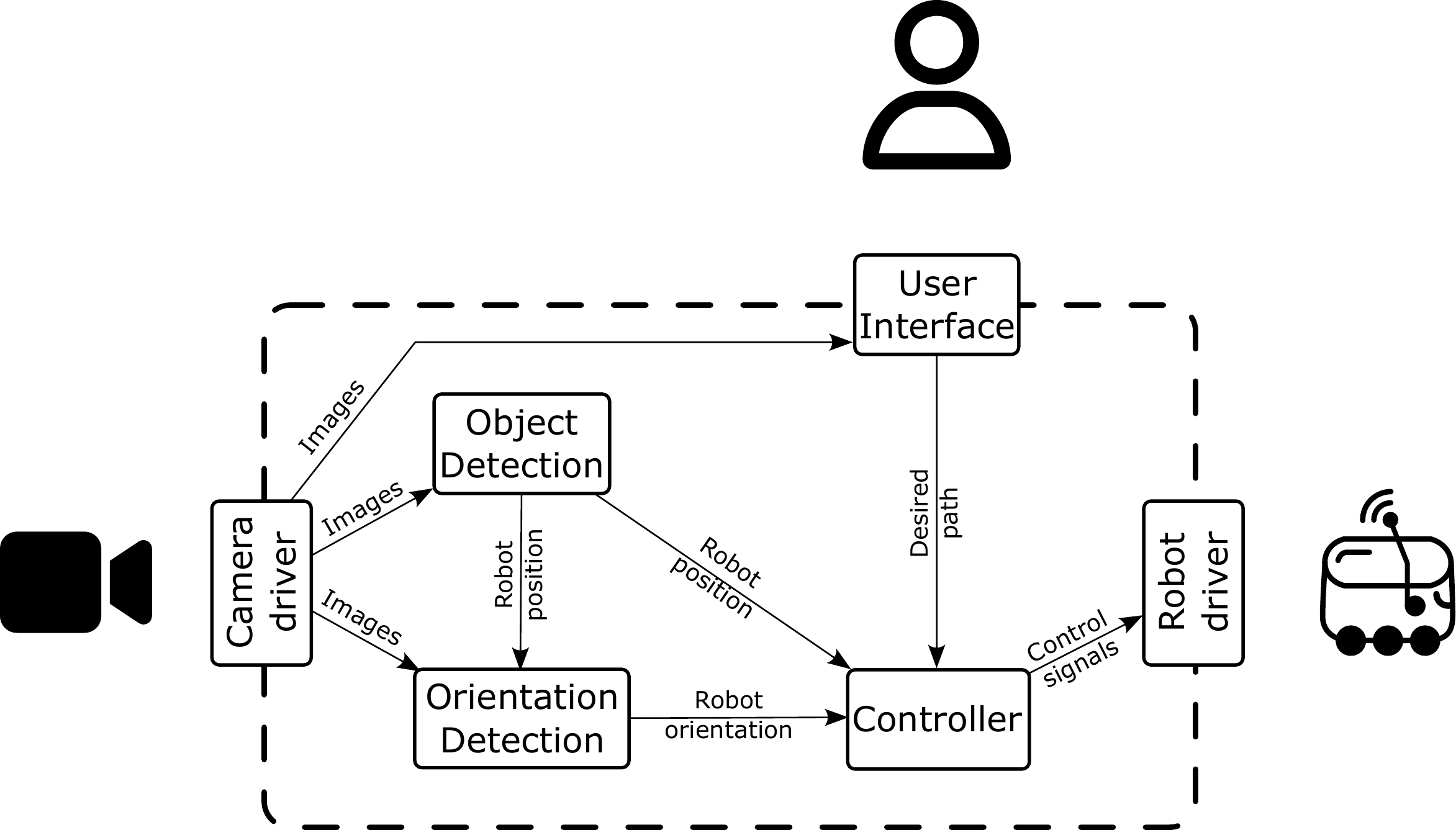}
\caption{Architecture of micro-services for the proposed visual servoing system.
The main computing unit -- dashed -- receives the cameras and converts them to control commands after estimating the position and the orientation of the robot.
The control signals are then sent wirelessly to the robot to follow a plan issued by a user through a GUI.}
\label{fig:system_diagram}
\end{figure}

\begin{figure*}[!h]
\centering
\includegraphics[width=\linewidth]{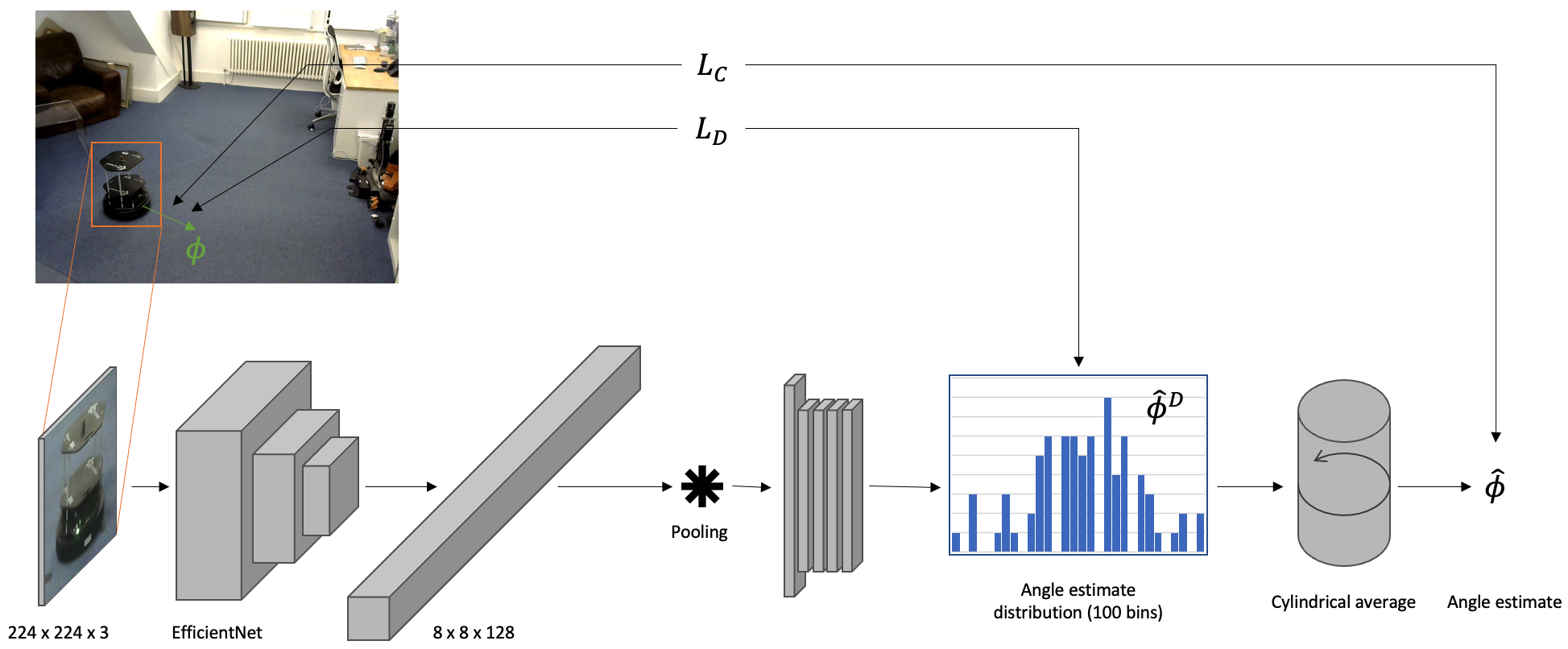}
\caption{Architectures of orientation detection.
A feature extractor down-scales the image to produce $8 \times 8$ feature vectors corresponding to the extracted area of the image containing the robot.
After pooling, it is passed through \gls{fc} layers to classify its rotation into a probability distribution over $100$ discrete angle bins before being converted in a single scalar angle estimate $\phi$ -- which is either continuous ($\phi^C$) or discretised ($\phi^D$) in separated parts of our positive loss $L_P=L_C+\alpha{}L_D$.}
\label{fig:orient_architecture}
\end{figure*}

\section{System Overview}%
\label{sec:overview}

Our in-image pose detection pipeline consists of two major parts: an object detector that finds robots in a video frame and marks them with bounding boxes and an orientation detector that takes each bounding box as its input and returns the orientation of the robot it contains.

Our core contribution is learning a \SI{360}{\degree} orientation-estimation model, described in~\cref{sec:orientation}.
Robust object detection is a prerequisite for orientation estimation, which we describe in~\cref{sec:yolo}.
Our contribution in this area is a robot-in-the-loop data-gathering, annotation, and training procedure which we describe in~\cref{sec:obj_det_training}.
\Cref{fig:system_diagram} depicts the overall micro-services architecture.

\subsection{Robot detection}
\label{sec:yolo}

For object detection, we chose YOLOv5~\cite{diwan2022object}\footnote{Our implementation is based on the code from Ultralytics \url{github.com/ultralytics/yolov5}}, due to the speed at which it can process image frames, various sizes of networks to choose from, and open-source implementation.
Critically for us, it and its predecessors are well-established as being successful in \textit{difficult, diverse real-world applications}.
Indeed, YOLO has been used in office spaces~\cite{dos2019mobile}, trained to recognise common indoor objects which were then used to inform a reactive control policy.
However, YOLO has also proven robust in diverse scenarios, having been used in traffic monitoring in urban scenes~\cite{lin2018yolo}.
Indeed, even in industrial applications, YOLO has been deployed in~\cite{xie2022package} to automate the task of counting packages entering and leaving a warehouse.
Finally, YOLO has in~\cite{li2022two} been used in automatic inspection for quality assurance.


\section{Orientation Estimation}%
\label{sec:orientation}

Our system, depicted in~\cref{fig:orient_architecture}, takes inspiration from \cite{ruiz2018fine}, to which we propose significant alterations in order to handle \SI{360}{\degree} rotations -- critical for our application, where the orientation of the robot is unrestrained.
Moreover, we switched the ResNet feature extractor for the more lightweight \textit{EfficientNet} model for its high performance and good scaling ability \cite{tan2019efficientnet}.
This is important in our application as we require lightweight performant devices at the edge.
Indeed, the ability of our network to scale down the number of parameters is very important since it allows the orientation detector to run faster, an important consideration given its central role in the control loop.

\cite{ruiz2018fine} is designed to regress a 3D rotation using one \gls{fc} classifier head for each angle.
As we only detect orientation in one dimension, we only need one \gls{fc} head after the \gls{cnn} for the one dimension.

Further, \cite{ruiz2018fine} limits the angles that the detector can sense to between \SI{-99}{\degree} and \SI{99}{\degree}.
For this reason, the expected value of the continuous angle can easily be regressed from the discrete probability distribution output of the classifier by summing the angles represented by each class multiplied by their probability.
However, once the limit on the angles to be detected is removed, this technique will no longer work due to the discontinuity of the angles wrapping back onto themselves at the transition from \SI{0}{\degree}-\SI{360}{\degree}.
To overcome this obstacle, a method from \cite{jammalamadaka2001topics} is used where the angle represented by each class, $\beta_i$, is first projected into two dimensions, $(x_i, y_i)$, where $x_i = \cos{\beta_i}$ and $y_i = \sin{\beta_i}$.
Then the expectation is taken separately in the two dimensions in the standard way, by summing the values multiplied by their probabilities.
This results in two expected values, $\bar{x}$ and $\bar{y}$.
These values can then be converted back into the expected angle using a quadrant aware $\arctan$ function, $\bar{\beta} = \atantwo(\bar{y}, \bar{x})$.
This method is fully differentiable and can be applied directly to the neural architecture.


Let $\phi$ and $\phi^D$ be the ground-truth value of the angle in image space, in its continuous and discretised form respectively.
Similarly, let $\hat{\phi}$ and $\hat{\phi}^D$ be the rotational estimates.

We can then define the \gls{mse} loss for sample $i$ between $\hat{\phi}_i$ and $\phi_i$ as:
\begin{equation}\label{eqn:LC}
\begin{split}
L_C = \sum_{i} \min \Bigl( ||\phi_i - \hat{\phi}_i||^2, & ||\phi_i + 2\pi - \hat{\phi}_i||^2, \\
&  ||\phi_i - 2\pi - \hat{\phi}_i||^2 \Bigr)
\end{split}
\end{equation}

Differently to~\cite{ruiz2018fine}, as our robots can be at any arbitrary orientation, to account for wrap-around errors, two additional \gls{mse} losses are calculated by adjusting $\hat\phi$ by adding or subtracting $2\pi$, and the minimum is selected.

Similarly, we can calculate the cross-entropy loss between the hot-encoded $\phi^D$ and $\hat{\phi}^D$ as
\begin{equation}
L_D = - \sum_{i} \sum_{j} \hat{\phi}^D_{i, j} \log \phi^D_{i, j}
\end{equation}
where $j$ ranges over the number of discretised bins.

Finally, we can combine $L_C$ and $L_D$ through a combination weight $\alpha$ to arrive at our total angular loss $L_A = L_C + \alpha L_D$.

\section{Fast-Deployment Robot-in-the-loop Training}%
\label{sec:pipeline}

We describe in this section the proposed procedure for data gathering and training the object and orientation detection networks -- all geared towards fast deployment.
As for the visual servoing paradigm, here we assume the camera position has a fixed viewpoint -- but we do not require knowledge of this fixed position.
Neither do we need information on the intrinsics of the camera, as all control will be done in image space.



\subsection{Spinning on the spot}
\label{sec:obj_det_training}

We manually drive the robot into five different locations in the image frame and observe it performing three \SI{360}{\degree} rotations without any translational movement.
This allows us to capture images of the robot's chassis from all sides and from different perspectives at different distances from the camera.
Critically, this is the only procedure that requires labels and is designed to produce labels very efficiently.
Indeed, we require human annotation for only the first and last frames at each spinning location and interpolate every frame in between, significantly reducing the amount of time and effort required for manual labelling.

To improve the robustness of our detection service, random brightness, contrast, image compression, and blur augmentations are applied to the training data.
Further, the input images to the network are normalised according to the mean and standard deviation of the \textit{ImageNet} dataset~\cite{deng2009imagenet}.

\subsection{Geofenced wandering}
\label{sec:orient_training}

To train the orientation estimator, the robot is programmed to navigate autonomously within a geofenced area, drawn in image space by the human operator through a user interface.
Upon reaching the area boundaries, the robot randomly spins before resuming straight driving until it reaches a boundary again.
The system uses the already-trained object detector to perform this task as well as a simple open loop rotation controller, calibrated from the manual spins.

Orientation labels are then obtained by measuring the bounding box location from the trained object detector in two frames that are only a short time apart.
The vector between these two bounding boxes provides an approximate but reliable estimate of the robot's orientation at the initial frame.
Using this process enables us to generate copious orientation labels efficiently in a short period of time and cover as much operational area as possible.
Labels are only generated for frames where the robot motion, measured by the in-image distance between two frames, is above a threshold.

We follow the same procedure described for the object detector in applying random augmentations and \textit{ImageNet} normalisation to the robot's bounding boxes before feeding them into the network for training.

\begin{figure}[t]
\centering
\begin{subfigure}{\columnwidth}
\centering
\includegraphics[width=0.8\columnwidth]{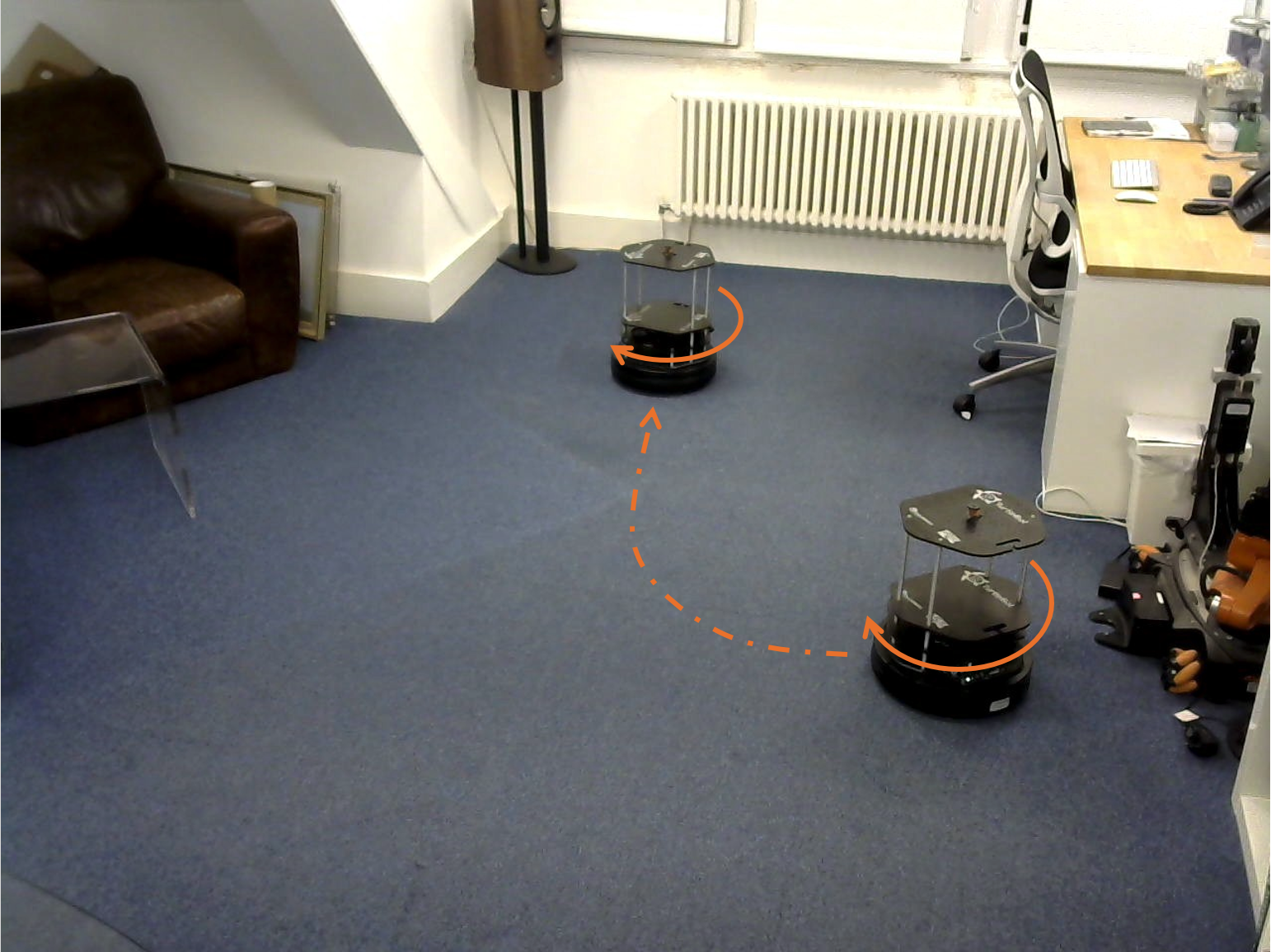}
\caption{}\label{fig:spinning}
\end{subfigure}

\begin{subfigure}{\columnwidth}
\centering
\includegraphics[width=0.8\columnwidth]{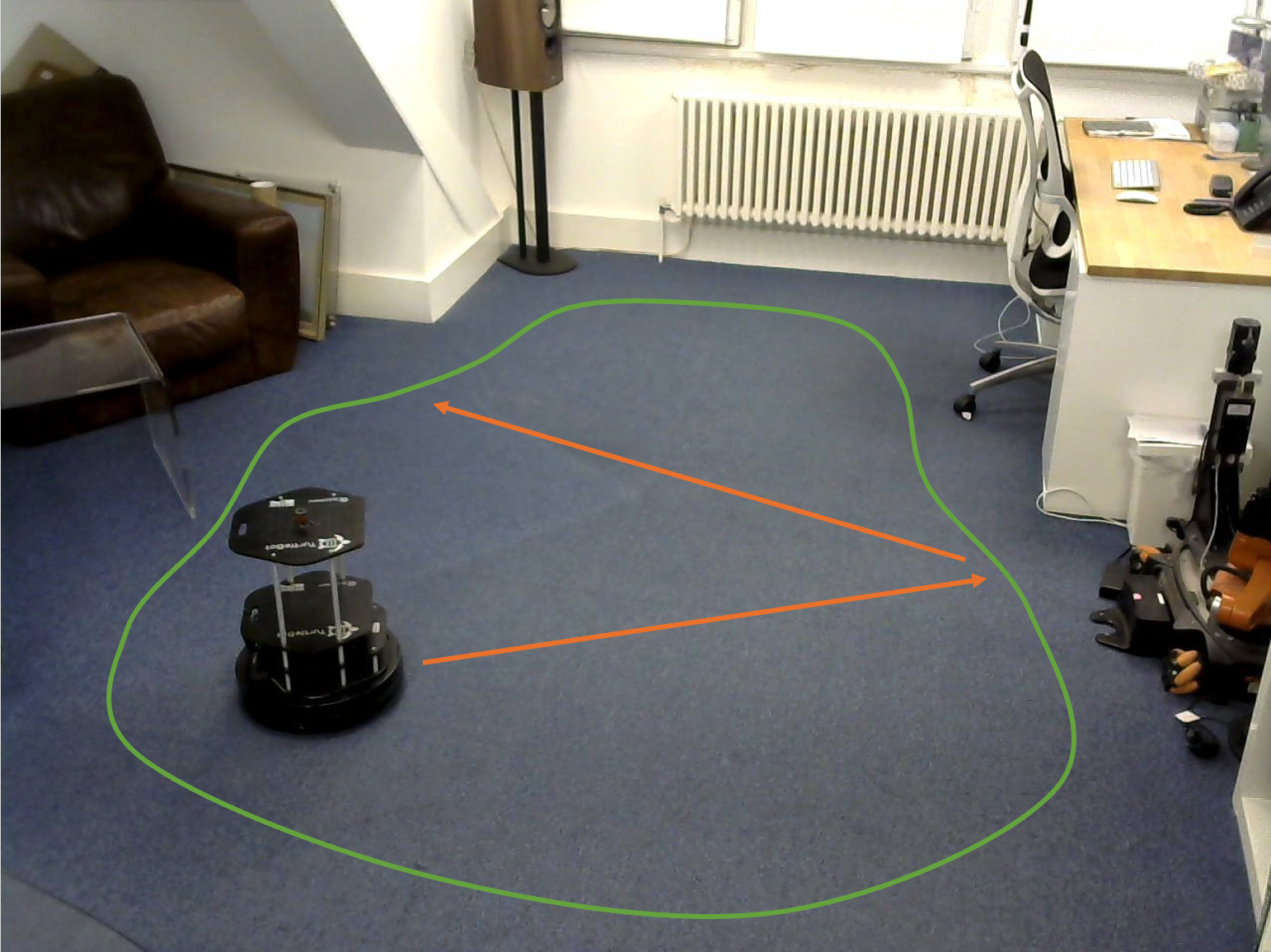}
\caption{}\label{fig:geofenced}
\end{subfigure}
\caption{
Capturing images of the robot in the presence of clutter and dynamic actors is important to train a robust object detection model that can handle real-world scenarios.
In such situations, the robot may be partially or fully occluded by other objects.
Furthermore, by capturing images under different lighting conditions, we can train an object detection model that is more robust to changes in illumination and can accurately detect the robot under varying lighting conditions.
\subref{fig:spinning} shows the object detection training procedure, where the robot is manually driven to five different locations to perform three \SI{360}{\degree} spins and \subref{fig:geofenced} shows the geofenced wandering used to train the orientation estimator.
}
\end{figure}

\section{Experimental Setup}%
\label{sec:experimental_setup}

We describe here the experimental setup for testing the performance of the proposed system.

\begin{figure*}
\centering

\begin{subfigure}{0.32\textwidth}
\includegraphics[width=\textwidth]{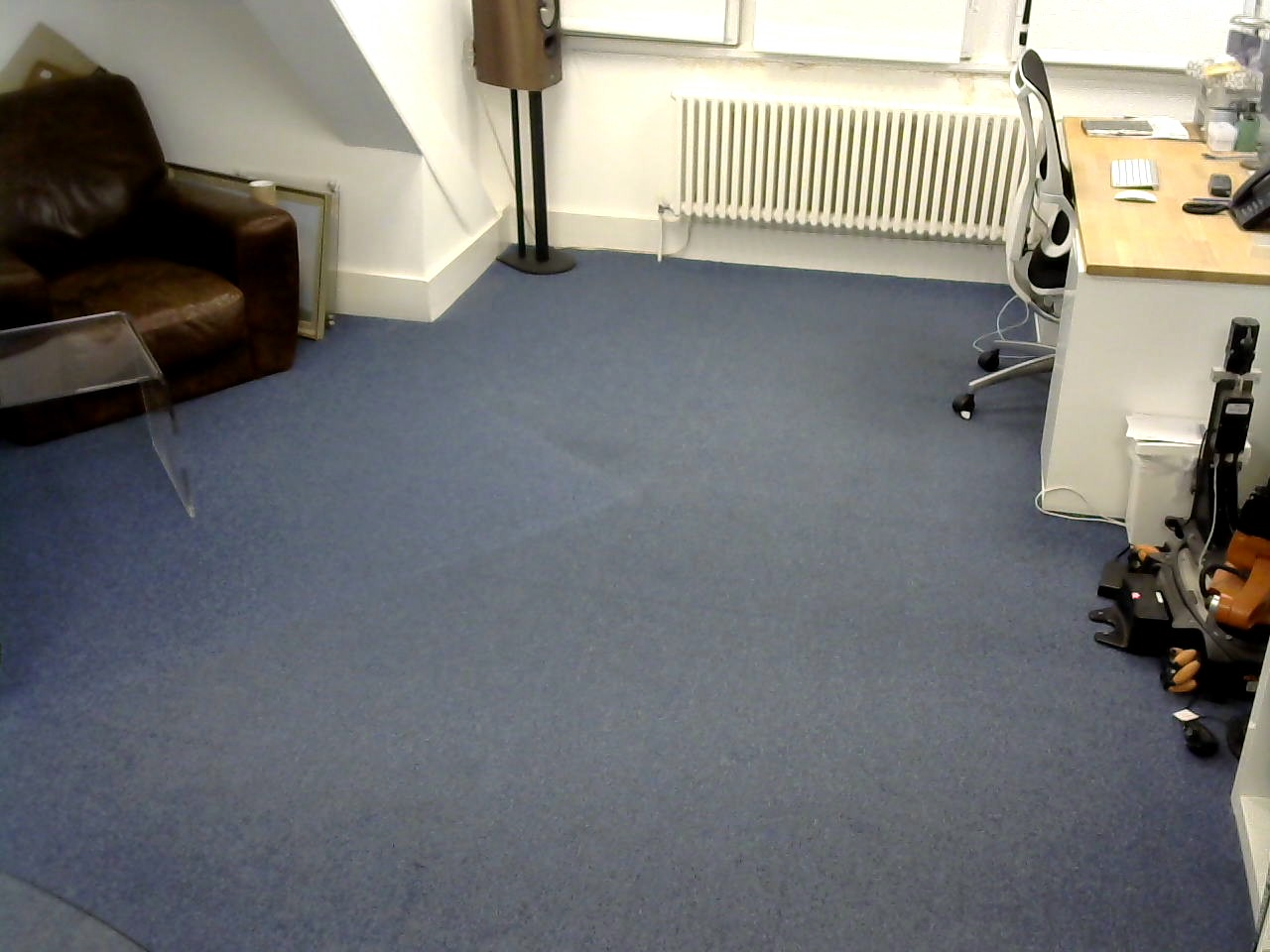}
\caption{\texttt{Office 1}}
\end{subfigure}
\begin{subfigure}{0.32\textwidth}
\includegraphics[width=\textwidth]{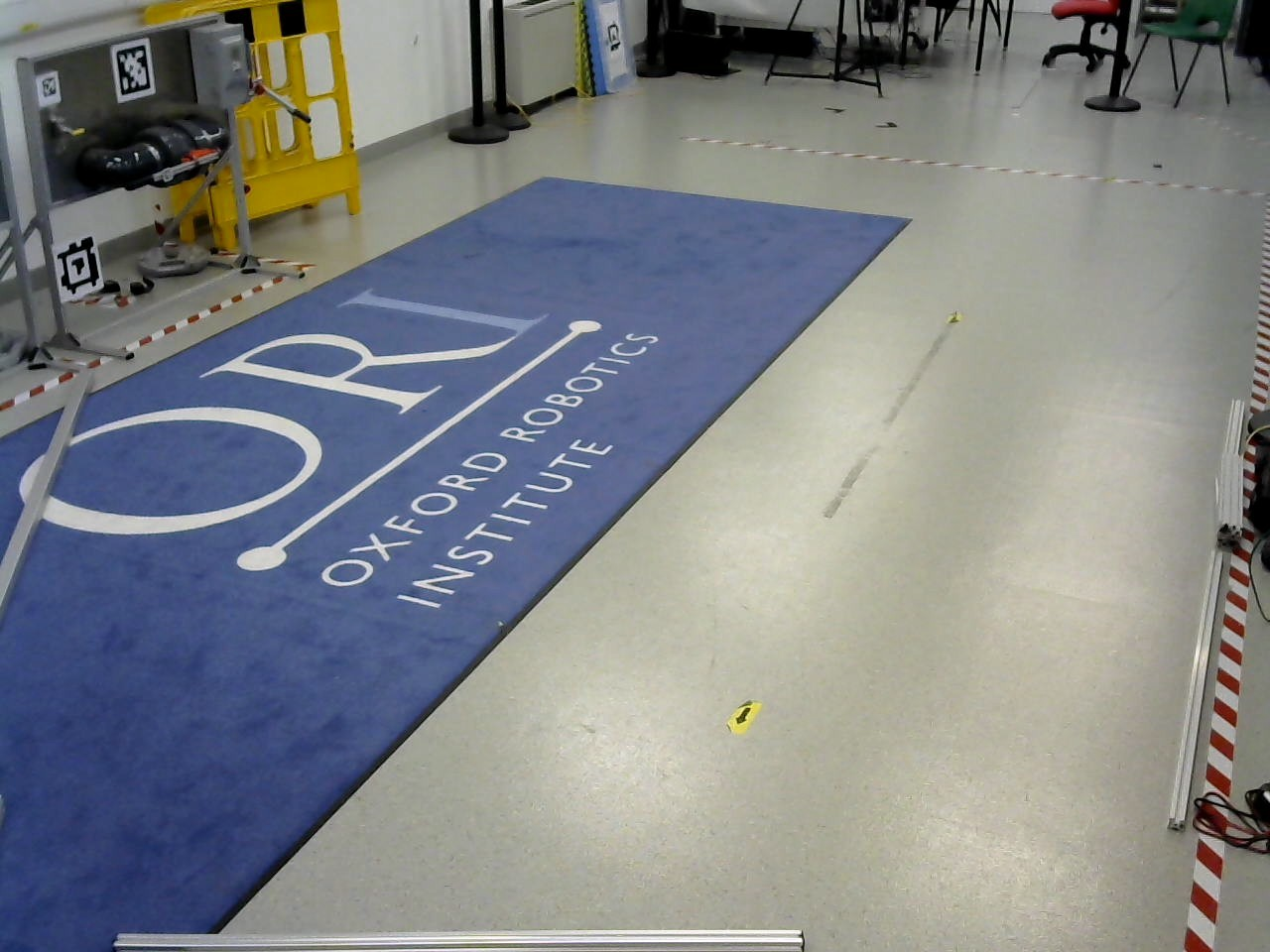}
\caption{\texttt{Workshop}}
\end{subfigure}
\begin{subfigure}{0.32\textwidth}
\includegraphics[width=\textwidth]{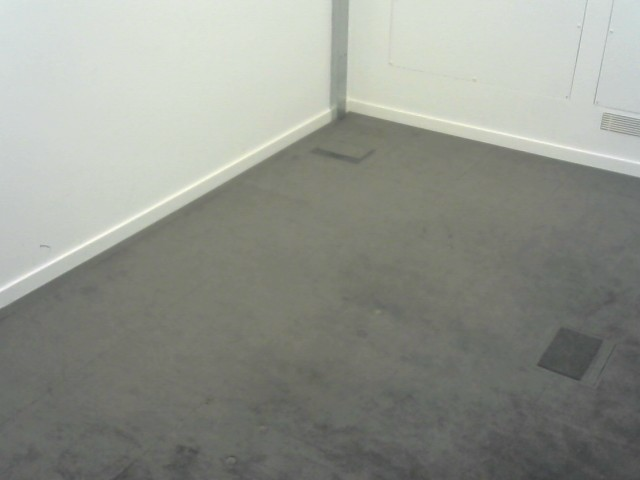}
\caption{\texttt{Empty Room}}
\end{subfigure}

\vspace{3pt}

\begin{subfigure}{0.32\textwidth}
\includegraphics[width=\textwidth]{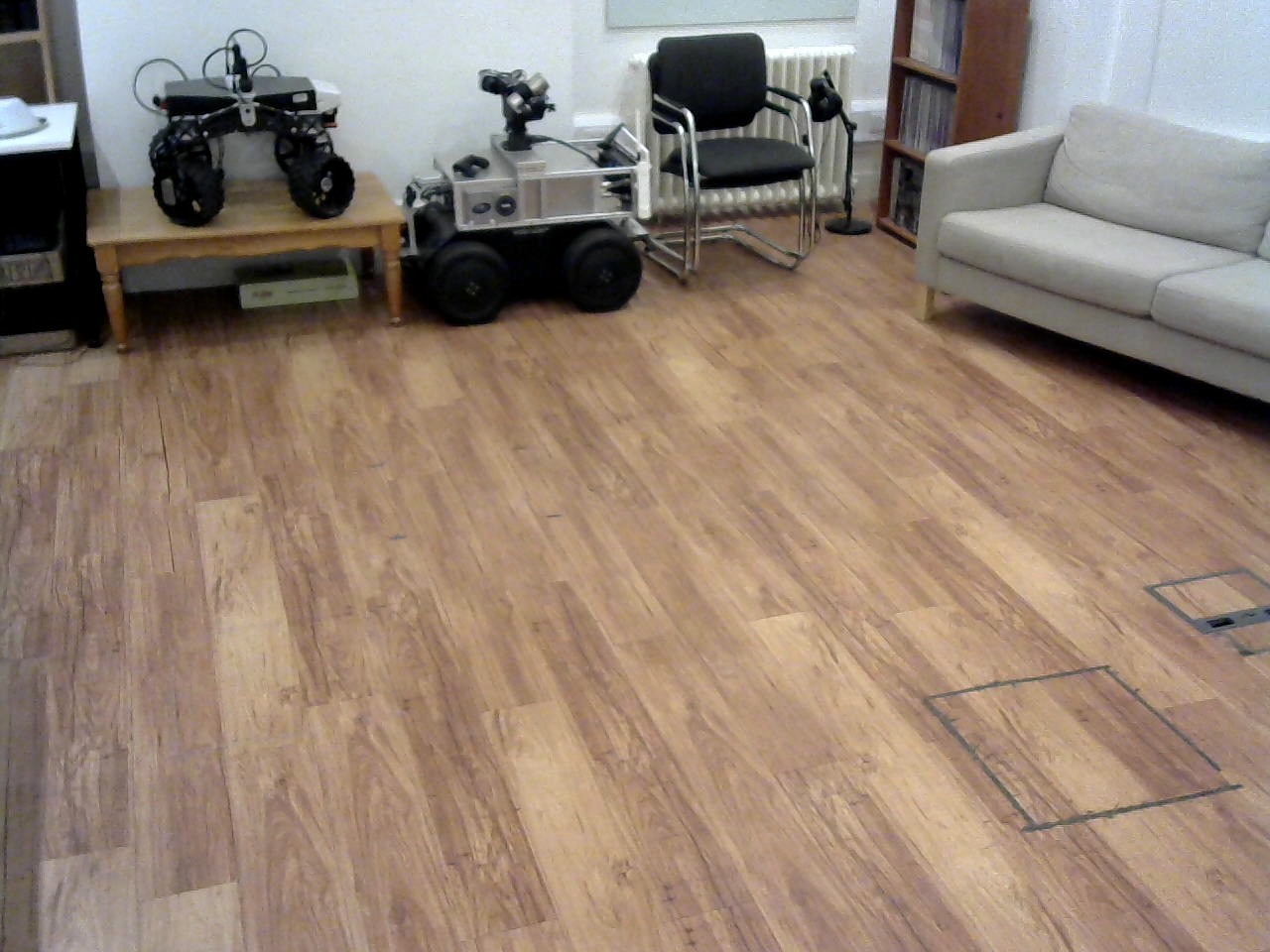}
\caption{\texttt{Meeting Room}}
\end{subfigure}
\begin{subfigure}{0.32\textwidth}
\includegraphics[width=\textwidth]{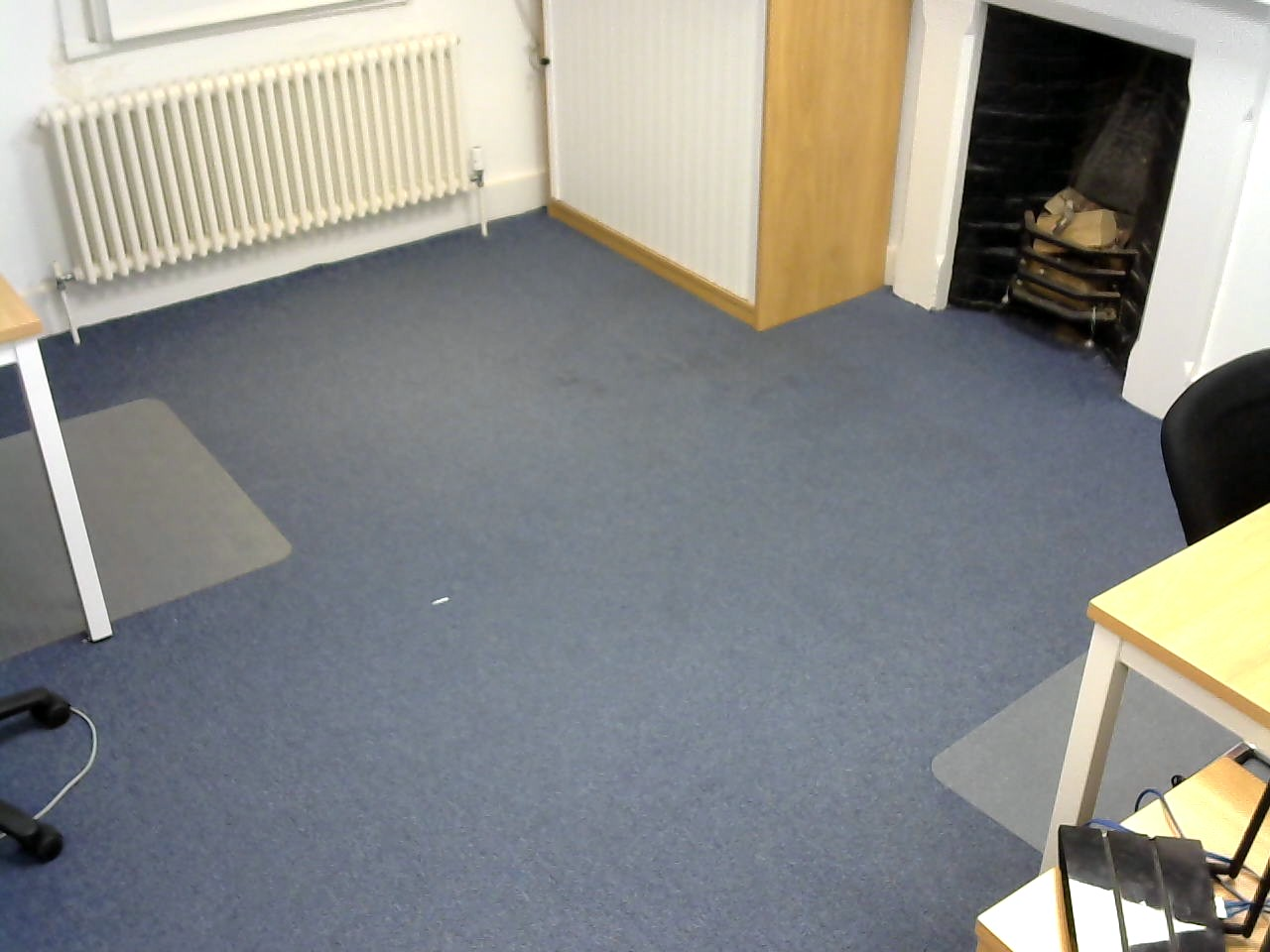}
\caption{\texttt{Office 2}}
\end{subfigure}

\caption{Images showing camera view for each location recorded; each location has different dimensions, textures, and typical objects to best evaluate our orientation detection pipeline.\label{fig:heatmaps}}
\end{figure*}

\subsection{Diverse scenarios and set-up}%
\label{sec:rooms}

We tested our proposed pipeline and neural architecture in five different indoor environments, as shown in \cref{fig:heatmaps}.
These have been chosen to have different textures, clutter, object types and dimensions to best evaluate the overall performance of the system.
In each of them, we install the camera -- a Logitech C270 HD Webcam -- on a tripod at approximately \SI{2.3}{\metre} from the floor.

The robot we employ is a Turtlebot2\footnote{\url{https://www.turtlebot.com/}}, to which we equip a Raspberry Pi 4 Model B driving the robot -- critically, the Raspberry Pi only acts as a message interpreter for actuator commands (forward speed and rotational speed) which are computed off-board and sent through WiFi.

The full autonomy stack -- from recording camera data to sending robot commands -- runs on a single laptop powered by an AMD Ryzen 9 5900HS CPU @ 3.30GHz with 16GB DDR4 memory and an NVIDIA GeForce RTX 3050 Ti GPU.
Training instead is carried out on a server equipped with an NVIDIA TITAN Xp and an NVIDIA GeForce GTX TITAN X.

We record \SI{45}{\minute} of geofenced wander for each environment to use as training, validation and test splits for our orientation estimator.
As for the test set, we reserve the last \SI{5}{\minute} of recording and for validation the last \SI{10}{\percent} of the training set, which varies in quantity depending on the following experiments.

\subsection{Repeatable Autonomous Traversals}%
\label{sec:noteachRepeat}
For each autonomous run, we specify a desired track the robot should follow.
For this, we mark the corners (i.e. turning points) of this track with tape on the ground.

To collect ground-truth positioning of the robot for use as a measure of performance, a Leica TS16 Laser Tracker\footnote{\url{https://leica-geosystems.com/products/total-stations/robotic-total-stations/leica-ts16}} tracker is installed in a fixed location within the room.
It tracks the precise movements of a prism which is attached to the top plate of the Turtlebot.
As the robot moves, the Leica tracker follows the prism and can determine the robot's position with millimetre accuracy in three dimensions.
However, it is important to note that the robot's movements are constrained to the ground plane and the Leica measurements are only used for analysing the result of the deployed robot, i.e. not used for robot control or training.

Then, we place the robot on each of these taped locations and take two measurements: one with the Leica to find the Leica-relative coordinates of the corner and one by marking a YOLO bounding box on the robot in the image to find the corresponding image coordinates of the corner.
Then, consider that the track comprises a series of line segments and that straight lines in three dimensions project to straight lines in the camera images (with lens distortion which is not too severe).
Therefore, we connect by straight line segments these bounding boxes and form an image-space path that the robot should follow, knowing that in doing so it will also follow on the ground plane the real-world line joining these waypoints.

\begin{figure*}[!h]
\centering
\begin{subfigure}[c]{\textwidth}
\begin{subfigure}[c]{0.48\textwidth}
\includegraphics[width=\textwidth]{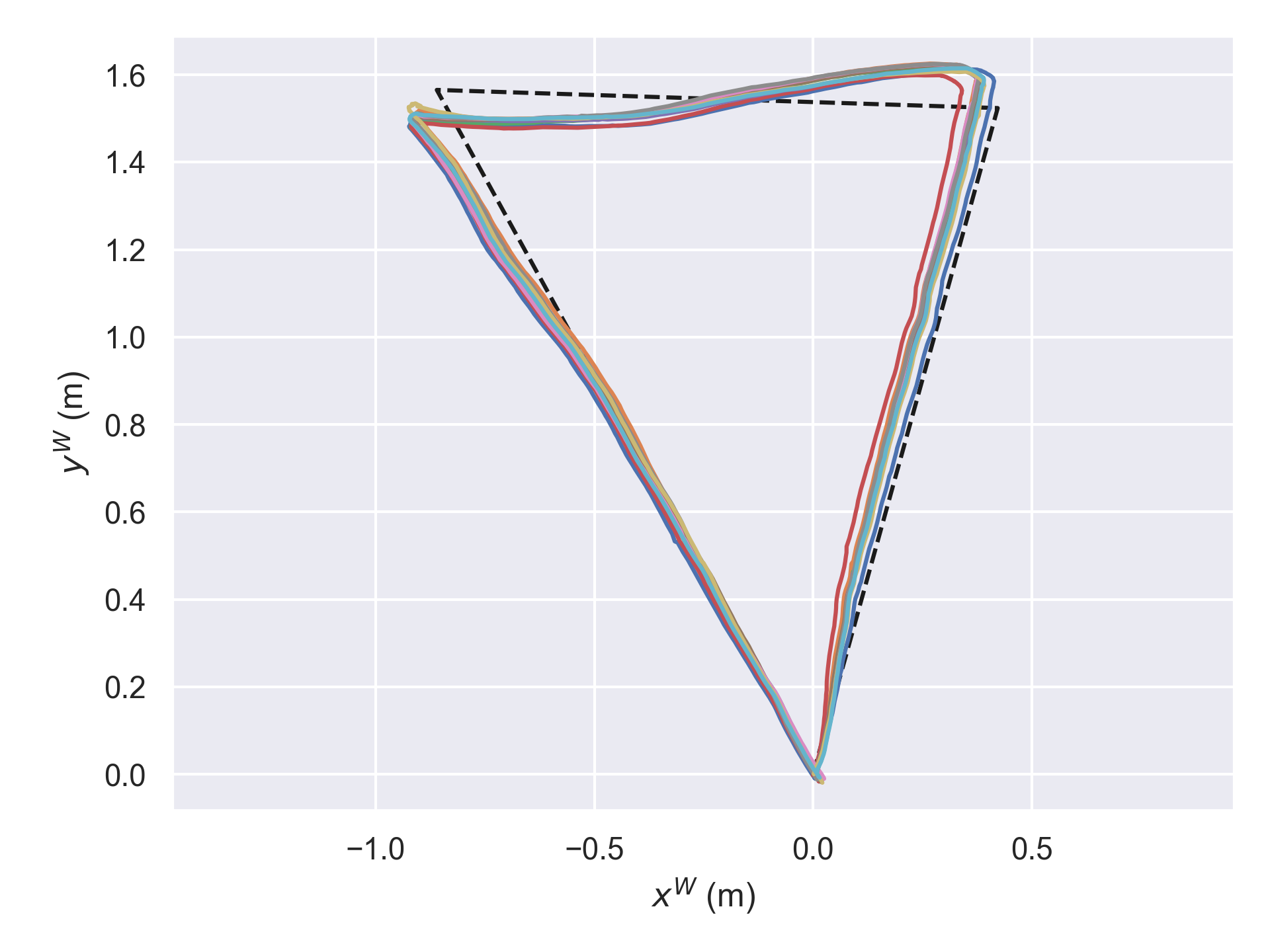}
\end{subfigure}
\begin{subfigure}[c]{0.48\textwidth}
\includegraphics[width=\textwidth]{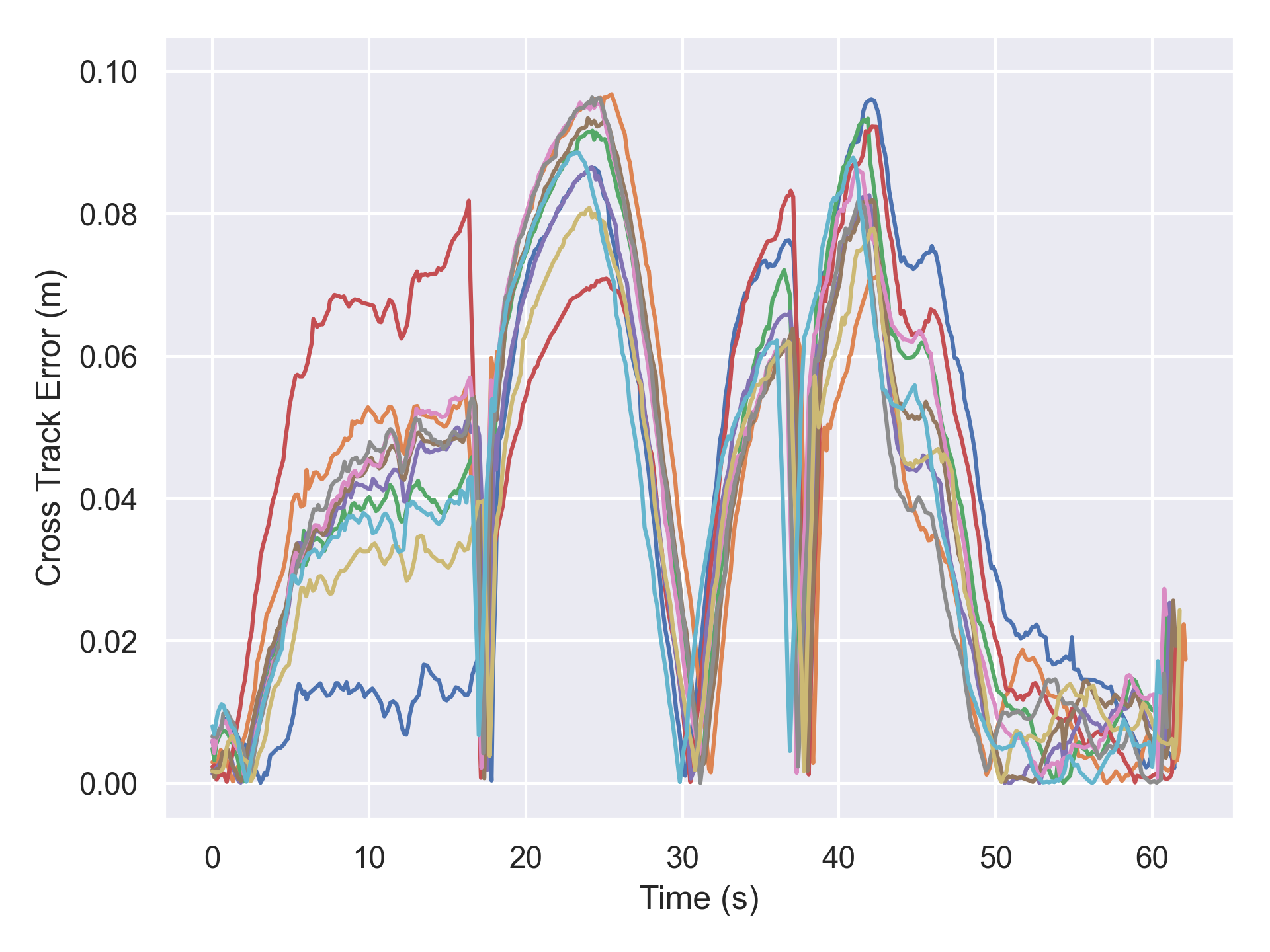}
\end{subfigure}
\caption{}
\end{subfigure}

\begin{subfigure}[c]{\textwidth}
\begin{subfigure}[c]{0.48\textwidth}
\includegraphics[width=\textwidth]{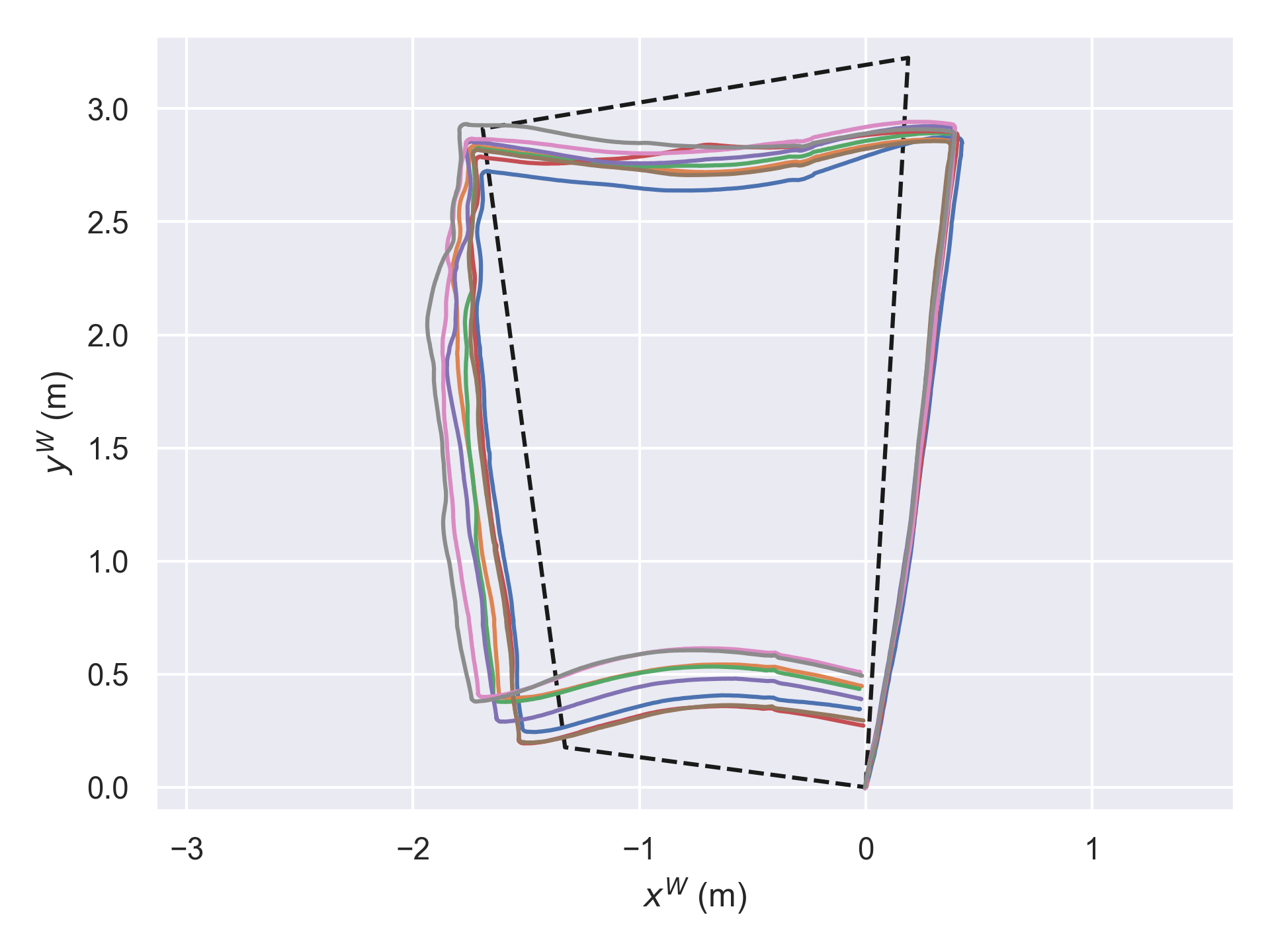}
\end{subfigure}
\begin{subfigure}[c]{0.48\textwidth}
\includegraphics[width=\textwidth]{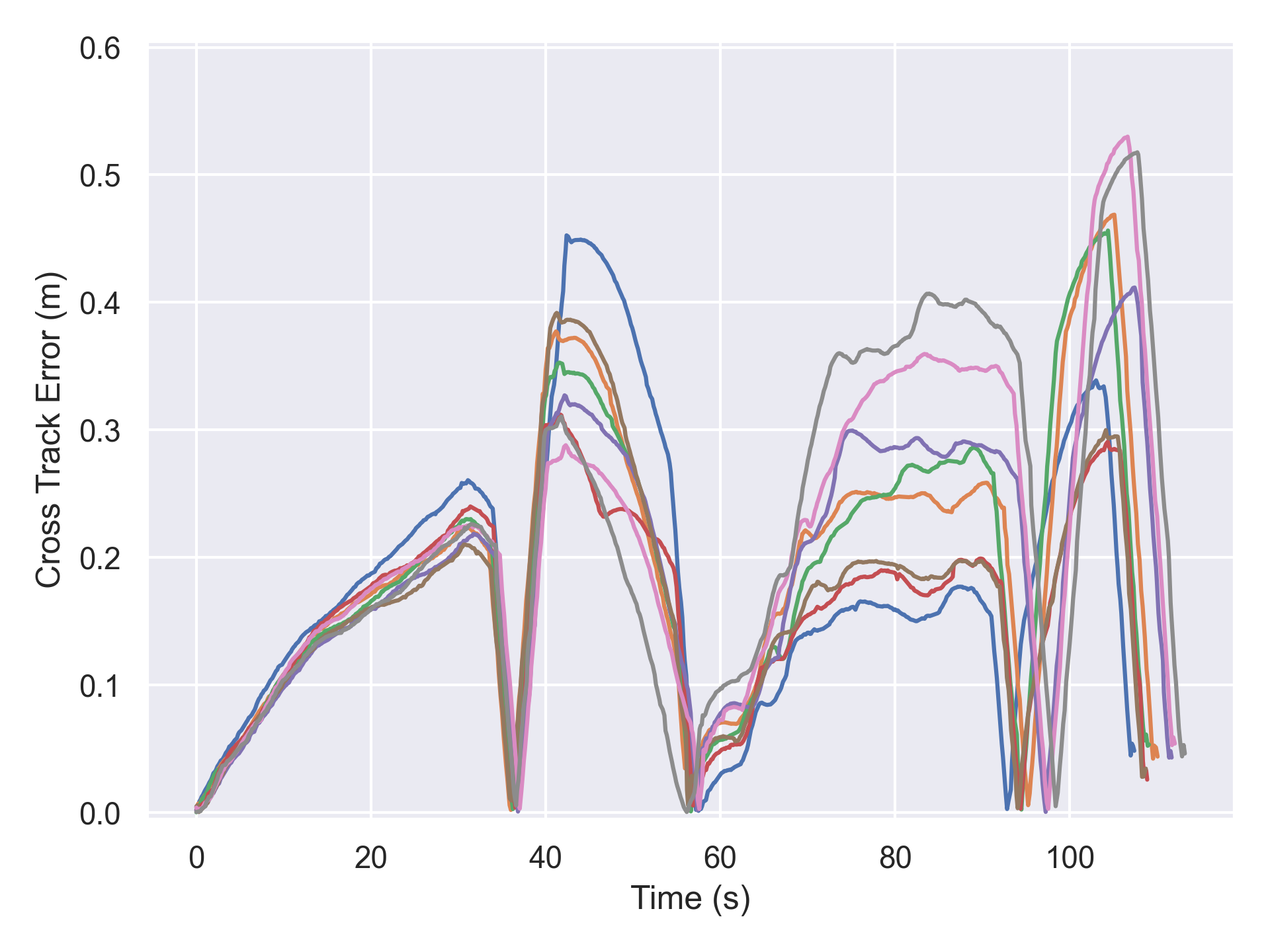}
\end{subfigure}
\caption{}
\end{subfigure}

\caption{
Full-autonomy test performed in \texttt{Workshop} (top) and \texttt{Office 1} (bottom).
(left) Tracks for the $8$ repeated autonomy runs in cartesian space. Here, the black dashed line is the desired track.
(right) Cross-track error vs time.
Here we can see how the robot has always been closer than \SI{10}{\centi\metre} from the track in \texttt{Office 1} and \SI{55}{\centi\metre} in \texttt{Workshop}. This discrepancy is probably due to the longer distance travelled by the robot in the workshop and the more horizontal point ov view, which challenges the in-image controller.
\label{fig:autonomy}}
\end{figure*}

A \gls{pd} controller takes as its input the pose of the robot \textit{in the image plane} and this intended track and outputs a linear and angular velocity command that is then sent to the robot, wirelessly.

\begin{table*}[!h]
\renewcommand{\arraystretch}{1.3}
\centering
\begin{subtable}{\textwidth}
\resizebox{\textwidth}{!}{
\begin{tabular}{c|cc|cc|cc|cc|cc}
& \multicolumn{2}{c|}{\texttt{Office 1}} & \multicolumn{2}{c|}{\texttt{Workshop}} & \multicolumn{2}{c|}{\texttt{Meeting Room}} & \multicolumn{2}{c|}{\texttt{Empty Room}} & \multicolumn{2}{c}{\texttt{Office 2}} \\
Model Architecture 
& median [\si{\degree}] & train time [\si{\minute}] & median [\si{\degree}] & train time [\si{\minute}] & median [\si{\degree}] & train time [\si{\minute}] & median [\si{\degree}] & train time [\si{\minute}] & median [\si{\degree}] & train time [\si{\minute}] \\
\hline
Pure Regression & \num{5.703238266} & \num{8.946} & \num{5.697402296} &\num{12.82} & \num{8.030581256} & \num{6.917} & \num{4.832763526} & \num{11.7} & \num{4.82764558} & \num{10.54} \\
Avg Pool - 1 FC & \num{2.771690783} & \num{1.982} & \num{2.900303241} & \num{2.599} & \num{2.762210493} & \num{2.68} & \num{2.163885873} & \num{2.144} & \num{2.42774120} & \num{2.802} \\
Max Pool - 1 FC & \num{3.247843775} & \num{2.255} & \num{3.359824673} & \num{2.685} & \num{3.212333569} & \num{2.678} & \num{2.518092658} & \num{2.139} & \num{2.78468990} & \num{2.772} \\
Avg Pool - 5 FC & \num{2.874679791} & \num{9.574} & \num{3.078770926} & \num{6.458} & \num{2.954275387} & \num{9.141} & \num{2.767477454} & \num{9.087} & \num{2.72670716} & \num{10.18} \\
Max Pool - 5 FC & \num{2.672065646} & \num{7.198} & \num{2.573806524} & \num{6.46} & \num{3.358839482} & \num{9.141} & \num{2.666174595} & \num{7.509} & \num{2.73659809} & \num{6.791} \\
\end{tabular}
}
\end{subtable}

\vspace{5pt}

\begin{subtable}{\textwidth}
\resizebox{\textwidth}{!}{
\begin{tabular}{c|cc|cc|cc|cc|cc}
& \multicolumn{2}{c|}{\texttt{Office 1}} & \multicolumn{2}{c|}{\texttt{Workshop}} & \multicolumn{2}{c|}{\texttt{Meeting Room}} & \multicolumn{2}{c|}{\texttt{Empty Room}} & \multicolumn{2}{c}{\texttt{Office 2}} \\
Model Architecture 
& median [\si{\degree}] & train time [\si{\minute}] & median [\si{\degree}] & train time [\si{\minute}] & median [\si{\degree}] & train time [\si{\minute}] & median [\si{\degree}] & train time [\si{\minute}] & median [\si{\degree}] & train time [\si{\minute}] \\
\hline
Pure Regression & \num{11.18158828} & \num{6.385} & \num{21.0369963} & \num{5.961} & \num{14.01785678} & \num{9.17} & \num{11.67603422} & \num{5.689} & \num{12.58948664} & \num{9.453} \\
Avg Pool - 1 FC & \num{6.405117416} & \num{2.167} & \num{9.843166495} & \num{0.476} & \num{9.75319596} & \num{1.828} & \num{5.905842614} & \num{1.726} & \num{7.597239918} & \num{1.509} \\
Max Pool - 1 FC & \num{7.405574563} & \num{2.000} & \num{29.33943586} & \num{0.2845} & \num{18.10836822} & \num{0.975} & \num{5.852502211} & \num{2.562} & \num{8.629042531} & \num{1.489} \\
Avg Pool - 5 FC & \num{5.880261856} & \num{5.205} & \num{9.400092404} & \num{1.151} & \num{11.5893282} & \num{5.198} & \num{5.138624537} & \num{5.880} & \num{4.935453715} & \num{2.714} \\
Max Pool - 5 FC & \num{5.976135438} & \num{3.220} & \num{7.431082071} & \num{2.97} & \num{7.976315237} & \num{3.473} & \num{4.920691937} & \num{2.478} & \num{6.859992691} & \num{4.429} \\
\end{tabular}
}
\end{subtable}

\caption{Accuracy and training-time values for the ablation study: (top) on full dataset and (bottom) on only the first \SI{5}{\minute}.}
\label{tab:model_ablation}
\end{table*}

\subsection{Measuring performance}%
\label{sec:crossTrack}

When measuring the performance of the orientation estimator alone, we produce ground truth in the same way as in~\cref{sec:orient_training} (box-to-box vectors).
We can thus report on median values of the errors between our predictions and this ground truth.

We thus have a 3D ground truth (piecewise line segments) by taking Leica positions at only three locations -- i.e. without performing the route manually first.
Under autonomous control (see~\cref{sec:noteachRepeat}) we record bounding box positions, and can therefore measure cross-track error with respect to this ground truth.
We report on aggregates of these cross-track errors over numerous autonomous runs.


\subsection{Architecture Search}

We performed an architecture search to select the best model for testing and deployment.
As shown in \cref{tab:model_ablation}, we tested different depths of the \gls{fc} head, as well as pooling operations and compared against a pure regression baseline.
Such baseline shares the architecture as the original model as in \cite{ruiz2018fine} but outputs a single scalar after a \texttt{tanh} activation.

We see all methods trained with both the discrete loss and the continuous loss outperforming the pure-regression model, both in terms of accuracy and training time.
A deeper \gls{fc} head does not result in a significant increase in accuracy when tested on the full dataset and is only visible when train with scarce data; nevertheless, it is negligible compared to the longer training times.
For this reason, from now on we will employ a model with average pooling and one single \gls{fc} layer for our experiments.

\section{Closed-Loop Autonomy Trials}%
\label{sec:openLoopResults}

In this section, we demonstrate the reliability of our orientation estimation pipeline over many diverse scenes.


Here we deployed the Turtlebot robot in two scenarios, \texttt{Workshop} and \texttt{Office 1}, in full autonomy using a \gls{pd} controller which computes cross-track and heading error with respect to a reference trajectory which is purely in image space.
A separate controller is used for following the line segments of the path and rotating on the spot between successive line segments with automatic switching between them based on robot pose.
\Cref{fig:autonomy} reports the cross-track error over $8$ consecutive runs at each location.

The system shows very high repeatability with a maximum of \SI{10}{\centi\metre} and \SI{55}{\centi\metre} error from the track as measured on the ground plane with the Leica Tracker.
Since the performance of pose detection is similar in both environments (\cref{tab:model_ablation}), it is thought that the increased maximum error of the \texttt{Workshop} autonomous runs is largely a result of the shallower angle of the camera with respect to the ground and the greater distance of the robot from the camera during its operation.

\section{Conclusion}%
\label{sec:conclusion}
We have presented a framework for deploying quickly and reliably a visual servoing application with no prior assumptions on either the scene or the robot -- e.g. visual markers.
We obtained these results by combining a robust off-the-shelf object detector with a novel orientation detection architecture and a robot-in-the-loop training procedure designed for minimising labelling effort and maximising area coverage.
We evaluated the proposed framework on indoor data from five different scenes of different sizes and with different clutter and textures.
Future works will focus on improving the controllability of the system in image space by taking into account the non-homogeneities due to the projective geometry of the scene.
Moreover, we would like to extend the proposed approach to a constellation of cameras, allowing wider coverage of areas.

\bibliographystyle{ieeetr}
\bibliography{biblio}

\end{document}